\documentclass[conference]{IEEEtran}

\IEEEoverridecommandlockouts

\usepackage[noadjust]{cite}
\usepackage{amsmath,amssymb,amsfonts}
\usepackage{algorithm}
\usepackage{algpseudocode}

\usepackage{dsfont}
\usepackage{graphicx}       
\usepackage{environ}         
\usepackage{etoolbox}        

\usepackage{textcomp}
\usepackage{setspace}
\usepackage[dvipsnames]{xcolor}
\usepackage{multirow}
\usepackage{caption}
\DeclareCaptionLabelFormat{algnonumber}{Algorithm}
\usepackage{hyperref}
\hypersetup{
    colorlinks=true,
    linkcolor=black,
    filecolor=magenta,      
    urlcolor=cyan,
    pdfpagemode=FullScreen,
    }

\usepackage{mathcomSTEv4}
\usepackage{spconf}

\usepackage[textfont=normalfont,singlelinecheck=off,justification=raggedright,font=footnotesize]{caption}

\usepackage[margin=30pt,textfont=normalfont,singlelinecheck=off,justification=raggedright,font=footnotesize]{subcaption}

\usepackage{anyfontsize}         

\usepackage{xcolor}
\usepackage{multicol}
\usepackage[utf8]{inputenc}

\usepackage{tikz,pgfplots}
\usetikzlibrary{arrows}

\newcommand{\randK}{ {\rm rand} K}
\newcommand{\topK}{ {\rm top} K}
\newcommand{\weightK}{ {\rm weighted} K}
\newcommand{\compk}{ {\rm comp}_K}
\newcommand{{\outk}}{ {\rm out}_K}

\newtheorem{remark}{Remark}

\usepackage{caption,booktabs}

\makeatletter
\def\endthebibliography{%
	\def\@noitemerr{\@latex@warning{Empty `thebibliography' environment}}%
	\endlist
}

\usepackage{glossaries}
\newacronym{memAOP}{Mem-AOP-GD}{Approximate Outer Product Gradient Descent with Memory}
\glsunsetall[main]

\title{
Speeding-up Back-propagation in DNN: \leavevmode \\
Approximate Outer Product with Memory 
}

\name{
Eduin E. Hernandez$^{\dagger}$ $\qquad$ Stefano Rini$^{\dagger}$ $\qquad$ Tolga M. Duman$^{\star}$
\vspace{-0.5cm}
}
\address{
$^{\dagger}$NYCU, Taiwan $\qquad$ $^{\star}$Bilkent University, Turkey \leavevmode \\
      $^{\dagger}$\{eduin.ee08, stefano.rini\}@nycu.edu.tw $\qquad$ $^{\star}$duman@ee.bilkent.edu.tr
      }
\begin{document}    
\maketitle
\begin{abstract}

In this paper, an algorithm  for approximate evaluation of back-propagation in DNN training is considered, which we term Approximate Outer Product Gradient Descent with Memory (\gls{memAOP}).
%
The \gls{memAOP} algorithm implements an approximation of the stochastic gradient descent by considering only a subset of the outer products involved in the matrix multiplications that encompass  back-propagation.
In order to correct for the inherent bias in this approximation, the algorithm retains in memory an accumulation of the outer products that are not used in the approximation.
We investigate the performance of the proposed algorithm in terms of DNN training loss under two design parameters: (i) the number of outer products used for the approximation, and (ii) the policy used to select such outer products.
We experimentally show that significant improvements in computational complexity as well as accuracy can indeed be obtained through \gls{memAOP}.
\end{abstract}


\section{Introduction}
Back-propagation is the cornerstone of Deep Neural Network (DNN) training \cite{hecht1992theory}.
Since its inception, many approaches have been considered to speed-up  back-propagation. As examples, adaptive learning rate (SuperSAB) \cite{silva1990speeding},  Newton-like methods (Quickprop) \cite{fahlman1988empirical},  second order methods \cite{battiti1992first} (conjugate gradient), an adaptive step-size method \cite{riedmiller1994advanced}  (RPROP), are some of the earliest and most celebrated algorithms to 
reduce its computational load. 
In this paper, we shall address the computational complexity of back-propagation by applying two sets of results for its computation: (i) approximate matrix multiplication, and (ii) error feedback in Stochastic Gradient Descent (SGD).
The resulting algorithm is named Approximate Outer Product Gradient Descent with Memory (\gls{memAOP}).
%
%
%
In \gls{memAOP}, we approximate the desired matrix product through a sum of outer products, in which the components are chosen from the original matrices according to some policy. 
Furthermore in order to correct for the error in the computation, we introduce a memory term, similar to \cite{stich2018sparsified}. 
The role of the memory is to account for the outer products that have not been selected for computation.
By judiciously combining the two ingredients, we show that the \gls{memAOP} provides improved performance over the classical back-propation algorithms.
Numerical evaluations with DNNs for regression and image classification are also presented to validate the proposed approach. 
%

{\bf Relevant Literature: }\label{sec:Relevant Literature}
\gls{memAOP}  combines three ideas: (i) approximate matrix multiplication, (ii) gradient compression, and (iii) SGD with memory. 
 Let us briefly introduce these concepts.

\noindent
{\bf Approximate matrix multiplication.}
Approximate matrix multiplication has a long history in mathematics, computer science, and engineering. 
The problem was initially considered in \cite{frieze2004fast}, inspired by the problem of finding low-rank matrix approximations in \cite{drineas2006fast}. 
When considering a simple algorithm in which the matrix multiplication is approximated by randomly sampling 
columns from $\Av$ and  rows from $\Bv$ (i.e., an outer product), and accumulating these rank-one  matrices to produce an approximation of $\Cv$, an approximation loss of $\Lcal(\Cv,\Chv)= \Ocal \lb \f{1}{\sqrt{c}} \| \Av \|_F \|\Bv \|_F \rb$
is obtained where $c$ is the number of outer products accumulated to produce $\Chv$ \cite{drineas2006fast}. 


\noindent
{\bf 
Error feedback in SGD.
}
It is widely accepted by DNN practitioners that gradient updates for DNN can be highly compressed without affecting the training performance.
In fact, for some cases compression of gradient updates improves the DNN performance.
Strategies for compressing gradients encompass sparsification and quantization, among others. 
%
%
For instance, $\topK$ is a rather aggressive sparsification method that keeps only the coordinates with the largest magnitudes \cite{alistarh2017qsgd,wangni2017gradient}.
%
%
When  DNN gradients are compressed, it has been shown that error feedback can greatly improve the classification/regression performance \cite{karimireddy2019error}. 
Error feedback for $1$-bit quantization was originally considered in \cite{seide20141}.
In \cite{stich2018sparsified}, error feedback is applied to gradient compression. 
Further results and generalization of this mechanism are provided in \cite{stich2019error}.


 

{\bf Organization:}
The paper is organized as follows: In  Sec. \ref{sec:Matrix Product}, we introduce relevant ingredients to the development of \gls{memAOP}.
In Sec. \ref{sec:Proposed Approach} the \gls{memAOP} is introduced.
We present  numerical experiments for both regression and classification tasks in Sec. \ref{sec:Numerical Experiment} and conclude the paper in Sec.  \ref{sec: conclusion}.

%
    
    
%
{\bf Notation:}\label{subsec:notations}
Lowercase boldface letters (e.g., $\zv$) are used for column vectors and uppercase boldface letters (e.g., $\Pv$) designate matrices.
We also adopt the shorthands $[m:n] \triangleq \{m, \ldots, n\}$
and  $[n] \triangleq \{1, \ldots, n\}$. 
%
Subscripts indicate the iteration index while $\Av^{(i)}$ and $\Av_{(i)}$ are used to denote the $i$th column and the $i$th row vector of matrix $\Av$, respectively.
%
%
%
The Frobenius norm of the matrix $\Av$ is indicated as $\| \Av \|_F$.
Calligraphic scripts are used to denote sets (e.g., $\Acal$) and $|\Acal|$ is used to denote its cardinality.
With $\Acal^c$ we indicate the complement of the set $\Acal$ when $\Omega$ is clear from the context.

\section{Relevant Results}
\label{sec:Matrix Product}

In this section, we shall first review back-propagation in DNNs, then  return to the techniques introduced in Sec. \ref{sec:Relevant Literature} and describe them in further detail.

\subsection{Back-propagation in DNNs}\label{subsec:Back-propagation for Deep Neural Networks}
Let us now denote the network as $D$ and the total number of layers as $I$. 
Let us further denote the weights, bias, inputs, outputs, and gradients of the corresponding layer as $\Wv_{i}$, $\bv_i$, $\Xv_{i}$, $\Ov_{i}$, and $\Gv_{i}$.
From this we can define the output of each layer as $\Ov_{i} = D_{i}(\Xv_{i})$, 
and the output of the network as $\Ov = D(\Xv)$,
where $\Xv = \Xv_{1}$ is the initial input and $\Ov = \Ov_{I}$ is the final output. 
Note that $\Ov_{i} = \Xv_{i+1}$. Then the forward propagation of a dense layer can be defined as
\ea{ \label{eq:foward_prop_eq}
    D_{i}(\Xv_{i})  &= \Xv_{i} \Wv_{i} + \bv_{i}.
}
For the back-propagation, we must consider two matrix products
\eas{ \label{eq:backprop_eq1}
    \Gv_{i} &=  \Gv_{i+1} \Wv_{i}^{T} \\
\label{eq:backprop_eq2}
    \Wv_{i}^{*} &=  \Xv_{i}^{T} \Gv_{i+1},
}{\label{eq:backprop}}
where \eqref{eq:backprop_eq1} is used for calculating the $\Gv_i$ used in $D_{i-1}$ for \eqref{eq:backprop_eq2} and $\Wv_{i}^{*}$ for updating $\Wv_{i}$, referred to as the gradient of the weight. 
If the optimizer is SGD, then  $\Wv_{i} = \Wv_{i} - \eta \Wv_{i}^{*}$.
For other optimizers (e.g., Adam \cite{kingma2014adam}), the steps for updating are according their respective rules.
Both \eqref{eq:backprop_eq1} and \eqref{eq:backprop_eq2} employ the chain rule to calculate their respective gradients. 
$\Gv_{L}$ is calculated from the derivative of the loss between $\Ov_{L}$ and $\Yv$, which is the ground truth. 
Note that for a single layer network, calculating \eqref{eq:backprop_eq1} is not necessary, as there is no $\Wv_{i-1}$ which requires an update. 


%
%

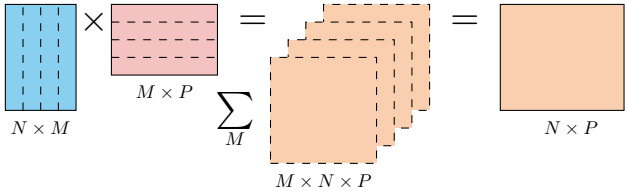
\begin{figure}[t]
 	\centering
	\definecolor{apricot}{rgb}{0.98, 0.81, 0.69}
\definecolor{antiquebrass}{rgb}{0.8, 0.58, 0.46}
\definecolor{arylideyellow}{rgb}{0.91, 0.84, 0.42}
\definecolor{bananamania}{rgb}{0.98, 0.91, 0.71}
\definecolor{babypink}{rgb}{0.96, 0.76, 0.76}
\definecolor{babyblue}{rgb}{0.54, 0.81, 0.94}
\definecolor{caribbeangreen}{rgb}{0.0, 0.8, 0.6}
\definecolor{celadon}{rgb}{0.67, 0.88, 0.69}
\definecolor{cornsilk}{rgb}{1.0, 0.97, 0.86}

\begin{tikzpicture}[scale=0.47, every node/.style={scale=0.7}]
\draw[fill=babyblue] (0,2) rectangle (2,5);
\draw[dashed] (0.5,2) -- (0.5,5);
\draw[dashed] (1.0,2) -- (1.0,5);
\draw[dashed] (1.5,2) -- (1.5,5);

\node[scale=2] at (2.5,4.5){$\times$};

\draw[fill=babypink] (3,5) rectangle (6,3);
\draw[dashed] (3, 4.5) -- (6, 4.5);
\draw[dashed] (3, 4.0) -- (6, 4.0);
\draw[dashed] (3, 3.5) -- (6, 3.5);
\node[scale=2] at (7,4.5) {$=$};

\draw[fill=apricot, dashed] (9,2) rectangle (12,5);
\draw[fill=apricot, dashed] (8.5,1.5) rectangle (11.5,4.5);
\draw[fill=apricot, dashed] (8,1) rectangle (11,4);
\draw[fill=apricot, dashed] (7.5,0.5) rectangle (10.5,3.5);
\node[scale=2] at (6.5,2) {$\sum$};
\node at (6.5, 1.2) {$M$};

\node[scale=2] at (13,4.5) {$=$};
\draw[fill=apricot] (14, 2) rectangle (17.5, 5);

\node at (1.0,1.5){$N \times M$};
\node at (4.5,2.5){$M \times P$};
\node at (9,0){$M \times N \times P$};
\node at (16,1.5){$N \times P$};

\end{tikzpicture}
    \vspace{-0.5cm}
    \caption{A conceptual representation of the outer product representation of the matrix multiplication.}
	\label{fig:outer_product}
    \vspace{-0.5cm}
\label{fig:}
\end{figure}

\subsection{Approximate Outer Product (AOP)}\label{subsec:Approximate Matrix Product}
One of the classical methods for performing matrix multiplication is through computing outer products for which sums of the outer products between the columns of $\Av$ and the rows of $\Bv$ are evaluated, i.e.,
\ea{\label{eq:outer_product}
\Cv = \textstyle \sum_{m \in [M]} \Av^{(m)} \Bv_{(m)},
}
where $\Av^{(m)} \in \mathbb{R}^{N \times 1}$ and $\Bv_{(m)} \in \mathbb{R}^{1 \times P}$.
The outer product operation is illustrated in Fig. \ref{fig:outer_product}.

In the context of DNN training, exact computations of matrix products is not always necessary; that is, obtaining and having approximations often suffices. 
%
Motivated by this observations, we can reduce the complexity of the training by sampling the columns of $\Av$ and rows of $\Bv$ for an Approximate Outer Product (AOP). 
%
Let us denote the sampling sets containing the indices of $[M]$ as $\Kcal$ and the size of this set as $|\Kcal|$ or $K$ for simplicity.
%

To obtain an approximation of the matrix product we choose $K < M$, and evaluate
\ea{\label{eq:approximated_outer_product}
\Chv = \textstyle \sum_{k \in \Kcal} \Av^{(k)} \Bv_{(k)},
}
where $\Chv$ is the resulting approximation of $\Cv$.
%
%
The elements in the set $\Kcal$ are most commonly chosen according to three strategies  (i) $\topK$, (ii) $\randK$, (iii) and $\weightK$.
In $\topK$, the elements with the largest 
values of $||\Av^{(k)}||_2 ||\Bv_{(k)}||_2$ are selected. 
 In $\randK$, the $K$ non-zero elements are  uniformly randomly selected from $[M]$.
Finally, in $\weightK$, the $K$ elements are selected at random with the probabilities
$p_k = {||\Av^{(k)}||_2 ||\Bv_{(k)}||_2}/{\sum ||\Av^{(k)}||_2||\Bv_{(k)}||_2}$.
Note that $\randK$ and $\weightK$ require a normalization constant\footnote{The following is necessary only if performed with random sampling with replacement.}, so that   \eqref{eq:approximated_outer_product} becomes
\ea{\label{eq:approximated_outer_product_with_constant}
\Chv = \textstyle \sum_{k \in \Kcal} \f{1}{p_{k}|\Kcal|} \Av^{(k)} \Bv_{(k)}.}
%
Note that $\Chv$ is an unbiased estimate of $\Cv$, i.e.,  $\mathbb{E} \Big[ \Chv \Big]  = \Cv $.
%
\subsection{Sparsified SGD with Memory}
In many scenarios of practical relevance, the communication, storage or aggregation of the stochastic gradients in SGD needs to be limited. 
In these scenarios, compression schemes are  often applied to the already calculated gradients. 
In gradient sparsification, the stochastic gradient in SGD is sparsified before being applied to the current model estimate. 
Various approaches have been proposed for sparification, mostly focusing on preserving the gradients with the largest magnitudes \cite{shi2019understanding}.
%

Since gradient compression reduces the amount of data, it decreases the learning the accuracy of the model and as consequence, may negatively affect the convergence to the optimal solution. 
%
As authors in \cite{stich2018sparsified} point out, using compression on the gradient does not guarantee convergence of the training. 
As such they introduce memory of the compressed gradient $m_{t}$ at time $t$, by defining
\eas{
    \Whv^*_t &\leftarrow \compk(m_t + \Wv^*_t) \\
    m_{t+1} & \leftarrow m_t + \Wv^*_t - \Whv^*_t,
}{\label{eq:memSGD}}
where $\compk( \ )$  is the gradient sparsification operator that sets all but the $K$ gradient entries to zero.
In \eqref{eq:memSGD}, $m_{t+1}$  is an error compensation term which accounts for the fact that the sparsified gradient $\Whv^*_t$ is applied to the model update, instead of  the full gradient $\Wv^*_t$.
\section{Proposed Approach: Mem-AOP-GD}
\label{sec:Proposed Approach}
In this section, we propose a combination of AOP and the SGD memory technique for calculating \eqref{eq:backprop_eq2}.
Instead of calculating the full gradient, at each layer $i$ and time $t$, \gls{memAOP} calculates an approximation of the gradient $\Whv^*_{i,t}$ by selecting a subset $\Kcal$ of outer products. 
Let $\outk$ be the function that selects the set of outer products.
%
To account for the matrix approximation error introduced by the operator $\outk$, and to help the SGD convergence, we will employ two memory matrices denoted as $m^X_t$ for $X$ and $m^G_t$ for $G$ that store columns and rows of the matrices not used in the previous computations.

The \gls{memAOP} algorithm for layer $i$ of the DNN is as follows:

\begin{algorithm}
\caption{AOP with Memory}\label{alg:AOP-with-Mem}
\begin{algorithmic}[1]
    \State{Initialize $\Wv_0$, $m^X_t$ = 0, $m^G_t$ = 0}
    \State{for $t \in [T]$, do}
	\State {\quad $\Xhv_t \leftarrow m^\Xv_t + \sqrt{\eta_t}\Xv_t$ }
	\State {\quad $\Ghv_t \leftarrow m^\Gv_t + \sqrt{\eta_t}\Gv_t$ }
    \State {\quad $\Kcal \leftarrow \outk(\Xhv_t, \Ghv_t)$}
    \State {\quad $\Whv^*_t \leftarrow \sum_{k \in \Kcal} \Xhv_t^{T,(k)} \Ghv_{t,(k)}$}
    \State {\quad $\Wv_{t+1} \leftarrow \Wv_t - \Whv^*_t$}
    \State {\quad $m^{\Xv}_{t+1,(k)} \leftarrow \Xhv_{t,(k)}, \ \forall \ k \in \Kcal^c$}
    \State {\quad $m^{\Gv}_{t+1,(k)} \leftarrow \Ghv_{t,(k)}, \ \forall \ k \in \Kcal^c$}
    \State {end for}
\end{algorithmic}
\caption{Pseudo code for the \gls{memAOP} algorithm}
\label{pseudo}
\end{algorithm}
In the above algorithm,  $\Wv_0$ are the initial weights at $t=0$, $\Xv_t$ and $\Gv_t$ are calculated as defined in Sec. \ref{subsec:Back-propagation for Deep Neural Networks}.
Note that, in the pseudocode, we do not explicitly indicate the index of the layer $i$ for simplicity:  the indexing according to the layer is as in Sec. \ref{subsec:Back-propagation for Deep Neural Networks}.
%
%
Also,  $t$ is the index of the current iteration and $T$ is used for the total number of iterations. 
The learning rate of the model is denoted by $\eta_t$ and the elements of the sets in $[M]$ not in $\Kcal$, is denoted as $\Kcal^c$. 

If one wishes to perform matrix approximation without memory, then the lines $8$ and $9$, which contain the update for the memory for both matrices, can be omitted.

\begin{remark} (DNN Optimizer)
Note that \gls{memAOP} is independent from the optimizer, since it only aids the approximate computation of the gradient weight. 
Since back-propagation is also used to compute the second moment of the gradient weights, \gls{memAOP} can also be applied to second order optimizers.
\end{remark}

Let us next provide a high-level perspective on why  \gls{memAOP} produces an efficient approximation of the DNN gradients. 
Let us consider the effect what is the results of the full matrix multiplication in line $6$ at time $t=2$ and for $\eta_t=1$ :
\ea{
\Whv^*_2&  = \lb m_2^{\Xv}+\Xv_2\rb^T \lb m_2^{\Gv}+\Gv_2\rb \nonumber \\
& = \Xv_2^T \Gv_2 + m_2^{\Xv,T}m_2^{\Gv} 
+ m_2^{\Xv, T} \Gv_2
+ \Xv_2^T m_2^{\Gv}.
\label{eq:res}
}{}
The terms in \eqref{eq:res} are: (i) $\Xv_2^T \Gv_2$ which is  desired gradient at time $t$: the optimizer  accumulate this term to the current model estimate, as in the classic SGD, 
(ii) $m_2^{\Xv,T}m_2^{\Gv}$ where $m_2^{\Xv,T}$/$m_2^{\Gv}$ is the matrix containing the column/rows of $\Xv_1^T$/$\Gv_1$ which were not evaluated at $t=1$.
By accumulating this term, the optimizer ``corrects'' the matrix approximation error of the previous iteration. 
(iii) $m_2^{\Xv,T} \Gv_2 + \Xv_2^T m_2^{\Gv}$ is a term which contains terms that would have not been accumulated through standard DNN training. 
At first glance, one might conclude that these terms constitute noise that only interferes with the training progress.
From variance reduction techniques for SGD, such as  SAGA \cite{defazio2014saga} and SVRG \cite{shang2018vr}, we know that stale gradients are still useful in the learning process.
We conjecture that, given the inherent memory in the gradient evolution, this latter term acts in a manner similar to a stale gradient, thus ultimately aiding the SGD convergence.
%
%
%
%



\section{Numerical Experiment}\label{sec:Numerical Experiment}

\begin{figure}[t]
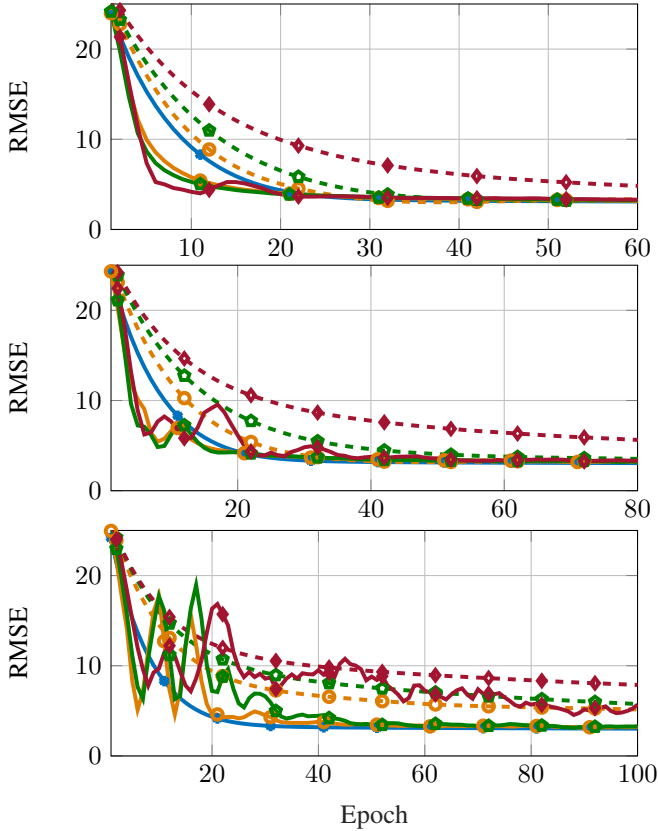

	\vspace{0.2cm}
	
	\begin{subfigure}[b]{0.5\textwidth}
		\input{tikz_files/Energy-Efficiency-18-over-144}
        \captionsetup{justification=centering}
	\end{subfigure}
	
	\begin{subfigure}[b]{0.5\textwidth}
		\input{tikz_files/Energy-Efficiency-9-over-144}
        \captionsetup{justification=centering}
	\end{subfigure}
	
	\begin{subfigure}[b]{0.5\textwidth}
		\input{tikz_files/Energy-Efficiency-3-over-144}
        \captionsetup{justification=centering}
	\end{subfigure}
	\vspace{-0.8cm}
	\caption{Comparison of 	
	validation loss for the the energy efficiency dataset  as discussed in Sec. \ref{sec:Numerical Experiment}. From top to bottom, $K=18,9,$ and $3$. In all, $M=144$.} \label{fig:EE_outer_product}
	\vspace{-0.5cm}
\end{figure}

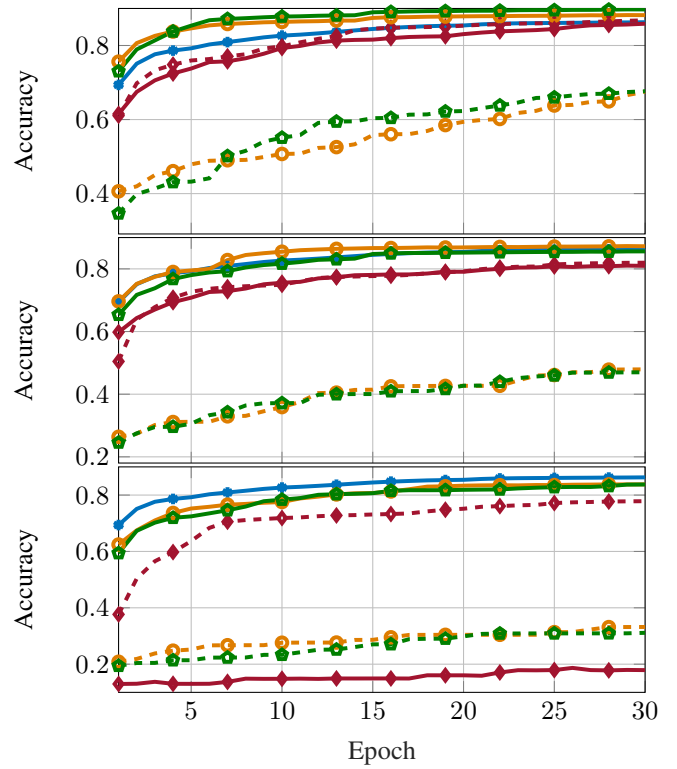
\begin{figure}[t]
	\vspace{0.2cm}
	
	\begin{subfigure}[b]{0.5\textwidth}
		\begin{tikzpicture}
\definecolor{mycolor1}{rgb}{0.00000,0.44706,0.74118}%
\definecolor{mycolor2}{rgb}{0.87059,0.49020,0.00000}%
\definecolor{mycolor3}{rgb}{0.00000,0.49804,0.00000}%
\definecolor{mycolor4}{rgb}{0.63529,0.07843,0.18431}%

\begin{axis}[%
width=7cm,
height=3cm,
scale only axis,
xmin=1,
xmax=30,
xlabel style={font=\color{white!15!black}},
xticklabels={,,}
ymin=0.3,
ymax=0.9,
ylabel={Accuracy},
axis background/.style={fill=white},
xmajorgrids,
ymajorgrids,
legend style={legend cell align=left, align=left, draw=white!15!black, nodes={scale=0.85, transform shape}, at={(0.01,0.65)}, anchor=west, fill opacity=0.8}
]

\addplot [color=mycolor1, line width=1.5pt, mark=asterisk, mark options={solid, mycolor1}, mark repeat={3}]
  table[row sep=crcr]{%
1	0.693653	\\
2	0.750983	\\
3	0.776263	\\
4	0.78629	\\
5	0.79235	\\
6	0.802803	\\
7	0.809203	\\
8	0.815433	\\
9	0.821977	\\
10	0.82665	\\
11	0.829303	\\
12	0.8328	\\
13	0.836697	\\
14	0.84068	\\
15	0.84462	\\
16	0.847653	\\
17	0.850043	\\
18	0.852103	\\
19	0.85329	\\
20	0.854113	\\
21	0.857167	\\
22	0.859073	\\
23	0.859967	\\
24	0.86048	\\
25	0.860857	\\
26	0.861307	\\
27	0.86162	\\
28	0.862133	\\
29	0.862367	\\
30	0.86286	\\
};

\addplot [color=mycolor2, line width=1.5pt, dashed, mark=o, mark options={solid, mycolor2}, mark repeat={3}]
  table[row sep=crcr]{%
1	0.40619	\\
2	0.41974	\\
3	0.45017	\\
4	0.46064	\\
5	0.47902	\\
6	0.48865	\\
7	0.49003	\\
8	0.49102	\\
9	0.49883	\\
10	0.50703	\\
11	0.50795	\\
12	0.52338	\\
13	0.52558	\\
14	0.53313	\\
15	0.55794	\\
16	0.56036	\\
17	0.56068	\\
18	0.56917	\\
19	0.58507	\\
20	0.59381	\\
21	0.59891	\\
22	0.60152	\\
23	0.61777	\\
24	0.62356	\\
25	0.63789	\\
26	0.63965	\\
27	0.64758	\\
28	0.64914	\\
29	0.66587	\\
30	0.67506	\\
};

\addplot [color=mycolor3, line width=1.5pt, dashed, mark=pentagon, mark options={solid, mycolor3}, mark repeat={3}]
  table[row sep=crcr]{%
1	0.34639	\\
2	0.3987	\\
3	0.41372	\\
4	0.43103	\\
5	0.43225	\\
6	0.44135	\\
7	0.50141	\\
8	0.51551	\\
9	0.5405	\\
10	0.55094	\\
11	0.55817	\\
12	0.59094	\\
13	0.59446	\\
14	0.5952	\\
15	0.60369	\\
16	0.60447	\\
17	0.61311	\\
18	0.61396	\\
19	0.62177	\\
20	0.62294	\\
21	0.63066	\\
22	0.63806	\\
23	0.64592	\\
24	0.65831	\\
25	0.65984	\\
26	0.66495	\\
27	0.66912	\\
28	0.66974	\\
29	0.67426	\\
30	0.67643	\\
};

\addplot [color=mycolor4, line width=1.5pt, dashed, mark=diamond, mark options={solid, mycolor4}, mark repeat={3}]
  table[row sep=crcr]{%
1	0.61056	\\
2	0.69793	\\
3	0.73308	\\
4	0.74824	\\
5	0.75952	\\
6	0.76343	\\
7	0.7706	\\
8	0.77632	\\
9	0.79175	\\
10	0.79773	\\
11	0.80853	\\
12	0.818	\\
13	0.82543	\\
14	0.84116	\\
15	0.84538	\\
16	0.84788	\\
17	0.84907	\\
18	0.84984	\\
19	0.85045	\\
20	0.85367	\\
21	0.85592	\\
22	0.85772	\\
23	0.85916	\\
24	0.85985	\\
25	0.86035	\\
26	0.8612	\\
27	0.8635	\\
28	0.86505	\\
29	0.86677	\\
30	0.86796	\\
};

\addplot [color=mycolor2, line width=1.5pt, mark=o, mark options={solid, mycolor2}, mark repeat={3}]
  table[row sep=crcr]{%
1	0.75637	\\
2	0.80656	\\
3	0.82604	\\
4	0.83767	\\
5	0.84621	\\
6	0.85435	\\
7	0.85809	\\
8	0.86021	\\
9	0.86258	\\
10	0.86372	\\
11	0.86496	\\
12	0.86568	\\
13	0.86728	\\
14	0.86725	\\
15	0.87448	\\
16	0.87635	\\
17	0.87661	\\
18	0.87764	\\
19	0.87812	\\
20	0.87874	\\
21	0.87919	\\
22	0.87923	\\
23	0.87992	\\
24	0.88017	\\
25	0.88038	\\
26	0.88103	\\
27	0.88161	\\
28	0.88138	\\
29	0.88217	\\
30	0.88196	\\
};

\addplot [color=mycolor3, line width=1.5pt, mark=pentagon, mark options={solid, mycolor3}, mark repeat={3}]
  table[row sep=crcr]{%
1	0.73039	\\
2	0.78941	\\
3	0.81371	\\
4	0.83781	\\
5	0.85358	\\
6	0.86825	\\
7	0.87169	\\
8	0.87378	\\
9	0.87617	\\
10	0.8781	\\
11	0.8789	\\
12	0.88033	\\
13	0.88116	\\
14	0.88165	\\
15	0.88968	\\
16	0.89067	\\
17	0.89187	\\
18	0.89276	\\
19	0.89285	\\
20	0.89367	\\
21	0.89392	\\
22	0.894	\\
23	0.89528	\\
24	0.89564	\\
25	0.89529	\\
26	0.89609	\\
27	0.89623	\\
28	0.89679	\\
29	0.89748	\\
30	0.89738	\\
};

\addplot [color=mycolor4, line width=1.5pt, mark=diamond, mark options={solid, mycolor4}, mark repeat={3}]
  table[row sep=crcr]{%
1	0.61513	\\
2	0.6748	\\
3	0.70584	\\
4	0.72442	\\
5	0.73777	\\
6	0.7552	\\
7	0.75721	\\
8	0.76544	\\
9	0.7778	\\
10	0.79321	\\
11	0.79882	\\
12	0.80815	\\
13	0.81318	\\
14	0.81505	\\
15	0.81583	\\
16	0.81997	\\
17	0.82298	\\
18	0.82416	\\
19	0.8248	\\
20	0.8307	\\
21	0.83487	\\
22	0.83812	\\
23	0.83985	\\
24	0.84162	\\
25	0.84443	\\
26	0.85021	\\
27	0.85444	\\
28	0.85542	\\
29	0.85666	\\
30	0.85877	\\
};

\end{axis}

\begin{axis}[%
width=7cm,
height=3cm,
at={(0in,0in)},
scale only axis,
xmin=0,
xmax=1,
ymin=0,
ymax=1,
axis line style={draw=none},
ticks=none,
axis x line*=bottom,
axis y line*=left
]

\end{axis}
\end{tikzpicture}%
        \captionsetup{justification=centering}
	\end{subfigure}
	
	\vspace{-0.2cm}
	\begin{subfigure}[b]{0.5\textwidth}
		\begin{tikzpicture}
\definecolor{mycolor1}{rgb}{0.00000,0.44706,0.74118}%
\definecolor{mycolor2}{rgb}{0.87059,0.49020,0.00000}%
\definecolor{mycolor3}{rgb}{0.00000,0.49804,0.00000}%
\definecolor{mycolor4}{rgb}{0.63529,0.07843,0.18431}%

\begin{axis}[%
width=7cm,
height=3cm,
scale only axis,
xmin=1,
xmax=30,
xlabel style={font=\color{white!15!black}},
xticklabels={,,}
ymin=0.2,
ymax=0.9,
ylabel={Accuracy},
axis background/.style={fill=white},
xmajorgrids,
ymajorgrids,
legend style={legend cell align=left, align=left, draw=white!15!black, nodes={scale=0.85, transform shape}, at={(0.01,0.65)}, anchor=west, fill opacity=0.8}
]

\addplot [color=mycolor1, line width=1.5pt, mark=asterisk, mark options={solid, mycolor1}, mark repeat={3}]
  table[row sep=crcr]{%
1	0.693653	\\
2	0.750983	\\
3	0.776263	\\
4	0.78629	\\
5	0.79235	\\
6	0.802803	\\
7	0.809203	\\
8	0.815433	\\
9	0.821977	\\
10	0.82665	\\
11	0.829303	\\
12	0.8328	\\
13	0.836697	\\
14	0.84068	\\
15	0.84462	\\
16	0.847653	\\
17	0.850043	\\
18	0.852103	\\
19	0.85329	\\
20	0.854113	\\
21	0.857167	\\
22	0.859073	\\
23	0.859967	\\
24	0.86048	\\
25	0.860857	\\
26	0.861307	\\
27	0.86162	\\
28	0.862133	\\
29	0.862367	\\
30	0.86286	\\
};

\addplot [color=mycolor2, line width=1.5pt, dashed, mark=o, mark options={solid, mycolor2}, mark repeat={3}]
  table[row sep=crcr]{%
1	0.26401	\\
2	0.27396	\\
3	0.30164	\\
4	0.31115	\\
5	0.31188	\\
6	0.31204	\\
7	0.32934	\\
8	0.33085	\\
9	0.34556	\\
10	0.35885	\\
11	0.37526	\\
12	0.40349	\\
13	0.40501	\\
14	0.41362	\\
15	0.41442	\\
16	0.42508	\\
17	0.42599	\\
18	0.4263	\\
19	0.42643	\\
20	0.42666	\\
21	0.42671	\\
22	0.42681	\\
23	0.44362	\\
24	0.45959	\\
25	0.4612	\\
26	0.46981	\\
27	0.47033	\\
28	0.47841	\\
29	0.47896	\\
30	0.47926	\\
};

\addplot [color=mycolor3, line width=1.5pt, dashed, mark=pentagon, mark options={solid, mycolor3}, mark repeat={3}]
  table[row sep=crcr]{%
1	0.24565	\\
2	0.27514	\\
3	0.2953	\\
4	0.29593	\\
5	0.30558	\\
6	0.33293	\\
7	0.34239	\\
8	0.36397	\\
9	0.37107	\\
10	0.37142	\\
11	0.3717	\\
12	0.39814	\\
13	0.39974	\\
14	0.40039	\\
15	0.40042	\\
16	0.40927	\\
17	0.40982	\\
18	0.40999	\\
19	0.41667	\\
20	0.42714	\\
21	0.42751	\\
22	0.43942	\\
23	0.45127	\\
24	0.45651	\\
25	0.45997	\\
26	0.4688	\\
27	0.46888	\\
28	0.4691	\\
29	0.46969	\\
30	0.4699	\\
};

\addplot [color=mycolor4, line width=1.5pt, dashed, mark=diamond, mark options={solid, mycolor4}, mark repeat={3}]
  table[row sep=crcr]{%
1	0.50458	\\
2	0.63828	\\
3	0.67809	\\
4	0.70782	\\
5	0.72686	\\
6	0.73646	\\
7	0.74148	\\
8	0.74466	\\
9	0.74691	\\
10	0.74918	\\
11	0.76035	\\
12	0.77083	\\
13	0.77422	\\
14	0.77557	\\
15	0.777	\\
16	0.77807	\\
17	0.78054	\\
18	0.78603	\\
19	0.78922	\\
20	0.79187	\\
21	0.79928	\\
22	0.80401	\\
23	0.80686	\\
24	0.80761	\\
25	0.81421	\\
26	0.81638	\\
27	0.81776	\\
28	0.81857	\\
29	0.81935	\\
30	0.81974	\\
};

\addplot [color=mycolor2, line width=1.5pt, mark=o, mark options={solid, mycolor2}, mark repeat={3}]
  table[row sep=crcr]{%
1	0.69635	\\
2	0.75238	\\
3	0.77473	\\
4	0.79017	\\
5	0.79497	\\
6	0.79979	\\
7	0.82815	\\
8	0.84429	\\
9	0.8501	\\
10	0.85474	\\
11	0.85961	\\
12	0.86192	\\
13	0.86397	\\
14	0.86505	\\
15	0.86643	\\
16	0.86654	\\
17	0.86742	\\
18	0.86864	\\
19	0.86817	\\
20	0.86855	\\
21	0.86931	\\
22	0.87003	\\
23	0.86968	\\
24	0.87036	\\
25	0.87111	\\
26	0.87116	\\
27	0.87152	\\
28	0.87191	\\
29	0.87282	\\
30	0.87221	\\
};

\addplot [color=mycolor3, line width=1.5pt, mark=pentagon, mark options={solid, mycolor3}, mark repeat={3}]
  table[row sep=crcr]{%
1	0.65345	\\
2	0.71706	\\
3	0.73821	\\
4	0.76768	\\
5	0.78019	\\
6	0.78856	\\
7	0.79269	\\
8	0.80454	\\
9	0.81229	\\
10	0.81655	\\
11	0.82188	\\
12	0.82886	\\
13	0.83057	\\
14	0.83263	\\
15	0.84756	\\
16	0.84936	\\
17	0.85075	\\
18	0.85078	\\
19	0.85145	\\
20	0.85187	\\
21	0.85244	\\
22	0.85287	\\
23	0.85345	\\
24	0.85385	\\
25	0.85414	\\
26	0.85431	\\
27	0.85497	\\
28	0.8548	\\
29	0.85573	\\
30	0.85584	\\
};

\addplot [color=mycolor4, line width=1.5pt, mark=diamond, mark options={solid, mycolor4}, mark repeat={3}]
  table[row sep=crcr]{%
1	0.59798	\\
2	0.64305	\\
3	0.67022	\\
4	0.6938	\\
5	0.70808	\\
6	0.72625	\\
7	0.7289	\\
8	0.73731	\\
9	0.75084	\\
10	0.75571	\\
11	0.75861	\\
12	0.76969	\\
13	0.7731	\\
14	0.77966	\\
15	0.78076	\\
16	0.78223	\\
17	0.78232	\\
18	0.78516	\\
19	0.79078	\\
20	0.79083	\\
21	0.79636	\\
22	0.80031	\\
23	0.80418	\\
24	0.80481	\\
25	0.80784	\\
26	0.80614	\\
27	0.80826	\\
28	0.80865	\\
29	0.81036	\\
30	0.80961	\\
};

\end{axis}

\begin{axis}[%
width=7cm,
height=3cm,
at={(0in,0in)},
scale only axis,
xmin=0,
xmax=1,
ymin=0,
ymax=1,
axis line style={draw=none},
ticks=none,
axis x line*=bottom,
axis y line*=left
]

\end{axis}
\end{tikzpicture}%
        \captionsetup{justification=centering}
	\end{subfigure}
	
	\vspace{-0.2cm}
	\begin{subfigure}[b]{0.5\textwidth}
		\begin{tikzpicture}
\definecolor{mycolor1}{rgb}{0.00000,0.44706,0.74118}%
\definecolor{mycolor2}{rgb}{0.87059,0.49020,0.00000}%
\definecolor{mycolor3}{rgb}{0.00000,0.49804,0.00000}%
\definecolor{mycolor4}{rgb}{0.63529,0.07843,0.18431}%

\begin{axis}[%
width=7cm,
height=3cm,
scale only axis,
xmin=1,
xmax=30,
xlabel style={font=\color{white!15!black}},
xlabel={Epoch},
ymin=0.1,
ymax=0.9,
ylabel={Accuracy},
axis background/.style={fill=white},
xmajorgrids,
ymajorgrids,
legend style={legend cell align=left, align=left, draw=white!15!black, nodes={scale=0.85, transform shape}, at={(0.01,0.65)}, anchor=west, fill opacity=0.8}
]

\addplot [color=mycolor1, line width=1.5pt, mark=asterisk, mark options={solid, mycolor1}, mark repeat={3}]
  table[row sep=crcr]{%
1	0.693653	\\
2	0.750983	\\
3	0.776263	\\
4	0.78629	\\
5	0.79235	\\
6	0.802803	\\
7	0.809203	\\
8	0.815433	\\
9	0.821977	\\
10	0.82665	\\
11	0.829303	\\
12	0.8328	\\
13	0.836697	\\
14	0.84068	\\
15	0.84462	\\
16	0.847653	\\
17	0.850043	\\
18	0.852103	\\
19	0.85329	\\
20	0.854113	\\
21	0.857167	\\
22	0.859073	\\
23	0.859967	\\
24	0.86048	\\
25	0.860857	\\
26	0.861307	\\
27	0.86162	\\
28	0.862133	\\
29	0.862367	\\
30	0.86286	\\
};

\addplot [color=mycolor2, line width=1.5pt, dashed, mark=o, mark options={solid, mycolor2}, mark repeat={3}]
  table[row sep=crcr]{%
1	0.20855	\\
2	0.21866	\\
3	0.23811	\\
4	0.24788	\\
5	0.25199	\\
6	0.26666	\\
7	0.26722	\\
8	0.26731	\\
9	0.26745	\\
10	0.27609	\\
11	0.27657	\\
12	0.27659	\\
13	0.27674	\\
14	0.28599	\\
15	0.28636	\\
16	0.29529	\\
17	0.30378	\\
18	0.30399	\\
19	0.30409	\\
20	0.30409	\\
21	0.30418	\\
22	0.30423	\\
23	0.30423	\\
24	0.30436	\\
25	0.31362	\\
26	0.31374	\\
27	0.32299	\\
28	0.33152	\\
29	0.33185	\\
30	0.33196	\\
};

\addplot [color=mycolor3, line width=1.5pt, dashed, mark=pentagon, mark options={solid, mycolor3}, mark repeat={3}]
  table[row sep=crcr]{%
1	0.1942	\\
2	0.20412	\\
3	0.20451	\\
4	0.21385	\\
5	0.21398	\\
6	0.22321	\\
7	0.22351	\\
8	0.22368	\\
9	0.23305	\\
10	0.23326	\\
11	0.24218	\\
12	0.2513	\\
13	0.25194	\\
14	0.26072	\\
15	0.2703	\\
16	0.27079	\\
17	0.28872	\\
18	0.29017	\\
19	0.2905	\\
20	0.29873	\\
21	0.3074	\\
22	0.30871	\\
23	0.30904	\\
24	0.30917	\\
25	0.30922	\\
26	0.30923	\\
27	0.30935	\\
28	0.30939	\\
29	0.30949	\\
30	0.31123	\\
};

\addplot [color=mycolor4, line width=1.5pt, dashed, mark=diamond, mark options={solid, mycolor4}, mark repeat={3}]
  table[row sep=crcr]{%
1	0.37668	\\
2	0.50367	\\
3	0.56529	\\
4	0.5977	\\
5	0.63686	\\
6	0.68355	\\
7	0.70564	\\
8	0.71138	\\
9	0.71519	\\
10	0.71852	\\
11	0.72071	\\
12	0.72507	\\
13	0.72778	\\
14	0.72895	\\
15	0.73118	\\
16	0.73273	\\
17	0.73482	\\
18	0.74218	\\
19	0.74798	\\
20	0.75151	\\
21	0.7585	\\
22	0.7608	\\
23	0.7636	\\
24	0.76517	\\
25	0.77164	\\
26	0.7742	\\
27	0.77546	\\
28	0.77706	\\
29	0.77775	\\
30	0.77827	\\
};

\addplot [color=mycolor2, line width=1.5pt, mark=o, mark options={solid, mycolor2}, mark repeat={3}]
  table[row sep=crcr]{%
1	0.62451	\\
2	0.67489	\\
3	0.70999	\\
4	0.73659	\\
5	0.75232	\\
6	0.75843	\\
7	0.76585	\\
8	0.76861	\\
9	0.77275	\\
10	0.77538	\\
11	0.78806	\\
12	0.79454	\\
13	0.80228	\\
14	0.80626	\\
15	0.80731	\\
16	0.81279	\\
17	0.8207	\\
18	0.82996	\\
19	0.83107	\\
20	0.83344	\\
21	0.83437	\\
22	0.8349	\\
23	0.83446	\\
24	0.83569	\\
25	0.83549	\\
26	0.83674	\\
27	0.8372	\\
28	0.83782	\\
29	0.83796	\\
30	0.838	\\
};

\addplot [color=mycolor3, line width=1.5pt, mark=pentagon, mark options={solid, mycolor3}, mark repeat={3}]
  table[row sep=crcr]{%
1	0.59397	\\
2	0.67315	\\
3	0.70214	\\
4	0.71933	\\
5	0.72447	\\
6	0.73517	\\
7	0.74608	\\
8	0.75928	\\
9	0.77821	\\
10	0.7837	\\
11	0.78934	\\
12	0.80092	\\
13	0.80317	\\
14	0.80453	\\
15	0.80663	\\
16	0.81511	\\
17	0.81727	\\
18	0.8172	\\
19	0.8174	\\
20	0.81865	\\
21	0.81938	\\
22	0.81984	\\
23	0.82363	\\
24	0.82609	\\
25	0.82803	\\
26	0.82883	\\
27	0.82874	\\
28	0.83187	\\
29	0.83708	\\
30	0.83756	\\
};

\addplot [color=mycolor4, line width=1.5pt, mark=diamond, mark options={solid, mycolor4}, mark repeat={3}]
  table[row sep=crcr]{%
1	0.12973	\\
2	0.13057	\\
3	0.13742	\\
4	0.13031	\\
5	0.13019	\\
6	0.13046	\\
7	0.13718	\\
8	0.14861	\\
9	0.14795	\\
10	0.14815	\\
11	0.14871	\\
12	0.14775	\\
13	0.14929	\\
14	0.14918	\\
15	0.14928	\\
16	0.14937	\\
17	0.14909	\\
18	0.16057	\\
19	0.16044	\\
20	0.16045	\\
21	0.15893	\\
22	0.17058	\\
23	0.17868	\\
24	0.17819	\\
25	0.17905	\\
26	0.18677	\\
27	0.179	\\
28	0.17799	\\
29	0.17957	\\
30	0.1789	\\
};

\end{axis}

\begin{axis}[%
width=7cm,
height=3cm,
at={(0in,0in)},
scale only axis,
xmin=0,
xmax=1,
ymin=0,
ymax=1,
axis line style={draw=none},
ticks=none,
axis x line*=bottom,
axis y line*=left
]

\end{axis}
\end{tikzpicture}%
        \captionsetup{justification=centering}
	\end{subfigure}
	\vspace{-0.8cm}
	\caption{Comparison of the validation loss for the MNIST dataset as discussed in Sec. \ref{sec:Numerical Experiment}. From top to bottom, $K=32,16,$ and $8$. In all, $M=64$.} \label{fig:MNIST_outer_product}
\end{figure}


In this section, we provide numerical experiments validating the performance of \gls{memAOP} it two cases:
(i) regression for the energy efficiency dataset \cite{tsanas2012accurate}, and 
(ii) classification for the MNIST dataset \cite{lecun2010mnist}.
In both cases, we investigate the performance the DNN training performance under two parameters\footnote{The code for the numerical experiment is available at  \url{https://github.com/HernandezEduin/Mem-AOP-GD}.}:
(i) the amount 
of computational reduction, and
(ii) the choice of the outer product selection operator.
%
The sampling is performed without replacement.
%

%
The simulation results are presented in Figs. \ref{fig:EE_outer_product} and \ref{fig:MNIST_outer_product}.
The legend for these figures is as follows: (i)  
blue:  baseline computation with matrix approximation,
(ii) yellow:   $\outk=\topK$, (iii) $\outk=\weightK$, and (iv)  red:  $\outk=\randK$.

In all the above cases, we also differentiate between two scenarios: (i) continuous line: \gls{memAOP} with memory and (ii) dashed line: \gls{memAOP} without memory. 


\smallskip

\noindent
$\bullet$ \underline{\bf Regression task:}
 In the dataset of \cite{tsanas2012accurate}, the DNN is tasked with assessing the heating load  of buildings as a function of building parameters. 
 %
The overall number of input features is $16$, after some pre-processing.
We train a simple single layered DNN with a weight of dimensions $16 \times 1$ and parameters detailed in the first column of Tab. \ref{tab:ML_parameters}. 
The effects of different levels of compression are shown in Fig. \ref{fig:EE_outer_product}.
Each row corresponds to an increasing rate: the lower the $K$, the higher the computational reduction level.

We observe that, for high values of $K$, \gls{memAOP} outperforms standard DNN back-propagation, despite the drastic reduction in its computational requirements. 
The performance of the different choices of $\outk$ with memory is rather close, although $\randK$ seems to yield a light improvement in performance. 
As the reduction in computation becomes more drastic, the effect of memory becomes less relevant.
%
Fig. \ref{fig:EE_outer_product} shows that memory in  \gls{memAOP}  might be disregarded when the amount of computational reduction is very large.
%

\smallskip

\noindent
$\bullet$\underline{\bf Classification task:}
%
Here we employ only a dense layer of $784 \times 10$ which come from the total number of pixel in an image and the  total numbers of digits to classify respectively. 
After the dense layer, we use a softmax activation layer. 
Further parameters are detailed in the second column of Tab. \ref{tab:ML_parameters}.
We choose a batch size of $64$ and approximate these batches with $K=32,16,$ and $8$ in the back-propagation from \eqref{eq:backprop_eq2}.
%
%
Note that for large values of $R=K/M$,  \gls{memAOP} yet again outperforms the standard back-propagation algorithm.
Also, as for the classification task, a drastic reduction in computational load does not significantly affect the training performance.
%
%
Interestingly, the performance of \gls{memAOP} with $\randK$ and without memory is rather competitive.
Inexplicably, the performance of \gls{memAOP} with $\randK$ and with memory falls drastically in performance for  the lowest value of $K$.

\begin{table}
    \vspace{-0.3cm}
	\footnotesize
	\centering
	\caption{Parameters and hyperparameters used for the training of the Machine Learning models.}
	\label{tab:ML_parameters}
	\begin{tabular}{|c|c|c|}
		\hline
		& Energy & MNIST \\ \hline
        Training Samples & 576 & 60k \\ \hline
        Validation Samples & 192 & 10k \\ \hline
        Optimizer & \multicolumn{2}{|c|}{SGD} \\ \hline
        Learning Rate & \multicolumn{2}{|c|}{0.01} \\ \hline
        Loss & MSE & Categorical Cross Entropy \\ \hline
        Epochs & 100 & 30 \\ \hline
        Mini-Batch Sizes & 144 & 64 \\ \hline
	\end{tabular}
	\vspace{-0.5cm}
\end{table}







\section{Conclusion} 
\label{sec: conclusion}
In this paper, the \gls{memAOP} algorithm is introduced for DNN training. 
This algorithms provides  a reduction in computational complexity together with an increase in training accuracy  by combining (i) matrix approximation through outer products and (ii) error feedback to obtain an approximate version of the back-propagation algorithm.
Numerical evaluations validate the performance of the proposed algorithm.
The theoretical analysis of \gls{memAOP} is left as a future research topic.
%

\bibliographystyle{IEEEtran}
\bibliography{bibs_matrix_mult}
\end{document}